\documentclass[10pt,twocolumn,letterpaper]{article}
\usepackage[pagenumbers]{cvpr}
\usepackage[T1]{fontenc}
\usepackage{colortbl}
\usepackage{tikz}
\usepackage{multirow}
\usepackage{microtype}
\definecolor{cvprblue}{rgb}{0.21,0.49,0.74}
\usepackage[pagebackref,breaklinks,colorlinks,allcolors=cvprblue]{hyperref}

\newcommand{\signaligner}{\textsc{SignAligner}\xspace}
\newcommand{\signalignerbf}{{\fontfamily{lmr}\fontseries{bx}\selectfont SignAligner}\xspace}
\newcommand{\scorebf}[1]{{\fontfamily{lmr}\fontseries{bx}\selectfont #1}}
\newcommand{\newpara}[1]{\vspace{1pt}\noindent{\fontfamily{lmr}\fontseries{bx}\selectfont #1}}
\newcommand{\sa}{\textsc{SA}\xspace}
\newcommand{\ssp}{\textsc{SS}\xspace}
\newcommand{\bobsl}{\textsc{BOBSL}\xspace}
\newcommand{\sentval}{\textsc{Sent-Val}\xspace}
\newcommand{\senttest}{\textsc{Sent-Test}\xspace}
\newcommand{\signtest}{\textsc{Sign-Test}\xspace}
\newcommand{\satplus}{\textsc{SAT}\textsuperscript{+}\xspace}

\newcommand{\PubLitfiePretrainingHours}{2{,}700}
\newcommand{\PubLitfieMultitaskMinusSsaFOneFifty}{+1.9}
\newcommand{\PubMomeniBaseSpotRecall}{25.4}
\newcommand{\PubMomeniBaseSpotCoverage}{13.9}
\newcommand{\PubMomeniExemplarsNovelSpotRecall}{45.6}
\newcommand{\PubMomeniExemplarsNovelSpotCoverage}{23.4}
\newcommand{\PubAudioFrameAccuracy}{40.1}
\newcommand{\PubAudioFOneTen}{46.4}
\newcommand{\PubAudioFOneTwentyFive}{33.5}
\newcommand{\PubAudioFOneFifty}{14.1}
\newcommand{\PubCorrectedAudioFrameAccuracy}{62.4}
\newcommand{\PubCorrectedAudioFOneTen}{72.8}
\newcommand{\PubCorrectedAudioFOneTwentyFive}{64.1}
\newcommand{\PubCorrectedAudioFOneFifty}{44.6}
\newcommand{\PubSatFrameAccuracy}{71.0}
\newcommand{\PubSatFOneTen}{74.2}
\newcommand{\PubSatFOneTwentyFive}{67.0}
\newcommand{\PubSatFOneFifty}{53.4}
\newcommand{\PubSatPlusFrameAccuracy}{77.2}
\newcommand{\PubSatPlusFOneTen}{81.4}
\newcommand{\PubSatPlusFOneTwentyFive}{75.0}
\newcommand{\PubSatPlusFOneFifty}{63.8}
\newcommand{\PubSeaBobslFOneFifty}{65.8}
\newcommand{\PubSeaSrfSatFOneFifty}{75.3}
\newcommand{\PubSeaSrfFOneFifty}{77.7}
\newcommand{\PubLitfieSsaFrameAccuracy}{75.7}
\newcommand{\PubLitfieSsaFOneTen}{83.7}
\newcommand{\PubLitfieSsaFOneTwentyFive}{79.7}
\newcommand{\PubLitfieSsaFOneFifty}{67.0}
\newcommand{\PubLitfieMultitaskFrameAccuracy}{77.1}
\newcommand{\PubLitfieMultitaskFOneTen}{84.4}
\newcommand{\PubLitfieMultitaskFOneTwentyFive}{80.3}
\newcommand{\PubLitfieMultitaskFOneFifty}{68.9}
\newcommand{\PubSeaSrfOriginalPlusFOneFifty}{74.8}

\newcommand{\SuGenerationsPerPrompt}{4}
\newcommand{\SuReleasedVerifiedSigns}{25{,}045}
\newcommand{\SuReleasedVerifiedClasses}{1{,}849}
\newcommand{\SuStageOneGPUHours}{157}
\newcommand{\SuHistoricalCentredSeedFOneFifty}{31.8}
\newcommand{\SuDistinctSignTestWords}{1{,}839}

\newcommand{\RfBOBSLFeatureEvaluatorFrameRateFps}{25}
\newcommand{\RfMaximumRefinementOptimiserSteps}{15{,}948}
\newcommand{\RfRefinementSeedMeanFOneZeroFifty}{66.68}
\newcommand{\RfSentTestEpisodeCount}{35}
\newcommand{\RfRefinementMeanBarY}{92}
\newcommand{\RfRefinementMinusMixedCEThreeSeedMeanFOneGainSigned}{+2.34}
\newcommand{\RfFilteredSentValSubtitleCount}{1{,}973}
\newcommand{\RfFilteredSentTestSubtitleCount}{20{,}338}
\newcommand{\RfApproximateGpuHoursPerRefinementSeed}{40}
\newcommand{\RfCorrectedRefinementSeedFortyTwoPooledFOne}{66.70}
\newcommand{\RfRefinementFOneTenMean}{84.67}
\newcommand{\RfRefinementFOneTwentyFiveMean}{79.67}
\newcommand{\RfRefinementFrameAccuracyMean}{77.37}
\newcommand{\RfMedianSignedStartOffsetSeconds}{-0.18}
\newcommand{\RfMedianSignedEndOffsetSeconds}{-0.06}
\newcommand{\RfDtwReconstructionRawSpanFOne}{64.72}
\newcommand{\RfEpisodeClusterBootstrapResamples}{10{,}000}
\newcommand{\RfHistoricalSaOnlyTestFOneZeroFifty}{62.9}
\newcommand{\RfHistoricalSatParentFOneAtOwnReconstruction}{64.7}
\newcommand{\RfDiagnosticRefinedSeedFOneAtOneDecimal}{66.7}
\newcommand{\RfFigureCaptionRefinementEpisodeWins}{32}
\newcommand{\RfCorrectedSeedFortyTwoGroundingRecall}{41.6}
\newcommand{\RfCorrectedSeedFortyTwoGroundingCoverage}{24.2}
\newcommand{\RfThresholdSweepEvaluatedSubtitles}{19{,}998}
\newcommand{\RfBootstrapPercentileIntervalLevel}{95}

\newcommand{\CtReselectedRefinementMinusMixedCeFOneGain}{+1.25}
\newcommand{\CtManualTrainingEpisodes}{16}
\newcommand{\CtManualSubtitleRows}{8{,}757}
\newcommand{\CtAuxiliaryWordRows}{7{,}198}
\newcommand{\CtLoraMillions}{12.2}
\newcommand{\CtMixedCeMean}{64.34}
\newcommand{\CtIouOnlyFOneFifty}{66.49}
\newcommand{\CtTotalParameterBillions}{3.24}
\newcommand{\CtStageOneTrainableMillions}{27.9}
\newcommand{\CtRefinementMinusParent}{+2.04}
\newcommand{\CtPilotOneMinusParent}{-0.77}
\newcommand{\CtPilotTwoMinusParent}{-0.09}
\newcommand{\CtMixedCeMinusParent}{-0.30}
\newcommand{\CtPilotOneFOneFifty}{63.87}
\newcommand{\CtPilotTwoFOneFifty}{64.55}
\newcommand{\CtSaOnlyCeMean}{64.48}
\newcommand{\CtParentContinuationPrompts}{0}
\newcommand{\CtParentFOneFifty}{64.64}
\newcommand{\CtParentDotY}{41}
\newcommand{\CtSaOnlyMaximumPrompts}{8{,}760}
\newcommand{\CtSaOnlyMinusParent}{-0.16}
\newcommand{\CtMixedMaximumPrompts}{63{,}840}
\newcommand{\CtPilotOneMaximumPrompts}{63{,}840}
\newcommand{\CtPilotTwoMaximumPrompts}{63{,}824}
\newcommand{\CtIouMaximumPrompts}{63{,}792}
\newcommand{\CtIouMinusParent}{+1.85}
\newcommand{\CtRefinementMaximumPrompts}{63{,}792}
\newcommand{\CtRefinementMinusPublishedSatPlus}{+2.9}
\newcommand{\CtRefinementMinusPublishedSea}{+0.9}

\newcommand{\JtFeatureGridFps}{12.5}
\newcommand{\JtWindowSeconds}{20}
\newcommand{\JtWindowPositions}{250}
\newcommand{\JtAudioPriorOffset}{2.7}
\newcommand{\JtSubtitleRows}{763{,}976}
\newcommand{\JtWordRows}{855{,}856}
\newcommand{\JtWordEpisodes}{1{,}658}
\newcommand{\JtWordConfidence}{0.9}
\newcommand{\JtTrainingEpochs}{5}
\newcommand{\JtQwenBudgetThousands}{400}
\newcommand{\JtOriginalJointFOne}{65.04}
\newcommand{\JtOriginalSubtitleFOne}{64.05}
\newcommand{\JtMaximumUpdates}{477{,}485}
\newcommand{\JtOriginalSubtitleDotY}{26}
\newcommand{\JtOriginalJointDotY}{51}
\newcommand{\JtOriginalGain}{+1.00}
\newcommand{\JtQwenSubtitleFOne}{63.7}
\newcommand{\JtQwenJointFOne}{64.1}
\newcommand{\JtDecoderfreeSubtitleWordPercent}{0}
\newcommand{\JtDecoderfreeWordPercentSeven}{7}
\newcommand{\JtDecoderfreeWordPercentFourteen}{14}
\newcommand{\JtDecoderfreeWordPercentTwentySix}{26}
\newcommand{\JtDecoderfreeWordPercentFortyOne}{41}
\newcommand{\JtDecoderfreeSubtitleFOne}{62.1}
\newcommand{\JtDecoderfreeSevenFOne}{61.2}
\newcommand{\JtDecoderfreeFourteenFOne}{60.0}
\newcommand{\JtDecoderfreeTwentySixFOne}{60.6}
\newcommand{\JtDecoderfreeFortyOneFOne}{59.4}
\newcommand{\JtDecoderfreeSmallestLoss}{0.89}
\newcommand{\JtDecoderfreeLargestLoss}{2.67}
\newcommand{\JtDecoderfreeSevenDelta}{-0.89}
\newcommand{\JtDecoderfreeFourteenDelta}{-2.07}
\newcommand{\JtDecoderfreeTwentySixDelta}{-1.45}
\newcommand{\JtDecoderfreeFortyOneDelta}{-2.67}
\newcommand{\JtBleurtReportingScale}{100}
\newcommand{\JtDecoderFreeTrainableMillions}{16.2}
\newcommand{\JtMrBeforeSaFOneFifty}{63.99}
\newcommand{\JtMrBeforeJointFOneFifty}{64.91}
\newcommand{\JtMrRefinedSaMeanFOneFifty}{65.42}
\newcommand{\JtMrRefinedJointMeanFOneFifty}{66.85}
\newcommand{\JtMrDBefore}{+0.92}
\newcommand{\JtMrMeanDAfter}{+1.43}

\newcommand{\WlFixedCentreSquareY}{28.90625}
\newcommand{\WlNativeMeanBarY}{86.23750}
\newcommand{\WlNativeMeanFOneHalf}{32.66}
\newcommand{\WlPositionControlMaximumFOneHalf}{3.15}
\newcommand{\WlNativeSeedFortyTwoFOneHalf}{32.61}
\newcommand{\WlWrongQueryFOneHalf}{4.59}
\newcommand{\WlFilteredSignTestQueries}{24{,}634}
\newcommand{\WlPositionControlLengthFrames}{20}
\newcommand{\WlFixedCentreFOneHalf}{3.15}
\newcommand{\WlUniformRandomMeanFOneHalf}{2.95}
\newcommand{\WlWholeWindowFOneHalf}{0.00}
\newcommand{\WlReferenceLengthSeconds}{0.8}
\newcommand{\WlSwsGroundingRecall}{39.9}
\newcommand{\WlSwsGroundingCoverage}{23.1}
\newcommand{\WlParentGroundingRecall}{40.9}
\newcommand{\WlParentGroundingCoverage}{23.8}
\newcommand{\WlSentCentreControlRecall}{98.5}
\newcommand{\WlSentCentreControlCoverage}{24.7}
\newcommand{\WlSignTestEpisodes}{250}
\newcommand{\WlCentredFixedCentreFOneHalf}{37.2}

\newcommand{\UtSentValidationEpisodes}{4}
\newcommand{\UtCombinedRealTimeFactor}{0.038}
\newcommand{\UtSignalignerSelectedCropCider}{59.1}
\newcommand{\UtSatplusSelectedCropCider}{58.8}
\newcommand{\UtSignalignerSelectedCropBleurt}{43.9}
\newcommand{\UtSatplusSelectedCropBleurt}{42.9}
\newcommand{\UtSatSelectedCropCider}{56.5}
\newcommand{\UtSatSelectedCropBleurt}{42.9}
\newcommand{\UtSrfrScratchMedianFOneFifty}{73.4}
\newcommand{\UtSrfrScratchRefinedFOneFifty}{73.0}
\newcommand{\UtSrfrScratchRefinedFOneTwentyFive}{88.5}
\newcommand{\UtSrfrScratchRefinedFOneTen}{93.2}
\newcommand{\UtSrfrParentFOneFifty}{76.6}
\newcommand{\UtSrfrParentFOneTwentyFive}{86.2}
\newcommand{\UtSrfrParentFOneTen}{87.7}
\newcommand{\UtSrfrParentRefinedFOneFifty}{77.1}
\newcommand{\UtSrfrParentRefinedFOneTwentyFive}{86.3}
\newcommand{\UtSrfrParentRefinedFOneTen}{87.6}
\newcommand{\UtSrfrFinalFOneFifty}{77.9}
\newcommand{\UtSrfrFinalFOneTwentyFive}{86.9}
\newcommand{\UtSrfrFinalFOneTen}{88.8}
\newcommand{\UtSrfrMatchedSaFOneFifty}{76.8}
\newcommand{\UtSrfrMatchedSaFOneTwentyFive}{84.2}
\newcommand{\UtSrfrMatchedSaFOneTen}{85.9}
\newcommand{\UtSrfrMatchedJointFOneFifty}{78.1}
\newcommand{\UtSrfrMatchedJointFOneTwentyFive}{86.1}
\newcommand{\UtSrfrMatchedJointFOneTen}{88.4}
\newcommand{\UtSrfrScratchMeanFOneFifty}{73.5}
\newcommand{\UtSrfrScratchMeanFOneTwentyFive}{87.7}
\newcommand{\UtSrfrScratchMeanFOneTen}{92.8}
\newcommand{\UtSrfrAudioFOneFifty}{46.9}
\newcommand{\UtSrfrAudioFOneTwentyFive}{75.6}
\newcommand{\UtSrfrAudioFOneTen}{84.8}
\newcommand{\UtSrfrTestSubtitles}{868}
\newcommand{\UtSrfrTestEpisodes}{4}
\newcommand{\UtSrfrScratchRefinedFrameAccuracy}{77.0}
\newcommand{\UtSrfrParentFrameAccuracy}{79.1}
\newcommand{\UtSrfrParentRefinedFrameAccuracy}{79.8}
\newcommand{\UtSrfrFinalFrameAccuracy}{79.7}
\newcommand{\UtSrfrMatchedSaFrameAccuracy}{79.7}
\newcommand{\UtSrfrMatchedJointFrameAccuracy}{80.7}
\newcommand{\UtSrfrScratchMeanFrameAccuracy}{78.3}
\newcommand{\UtSrfrAudioFrameAccuracy}{64.1}
\newcommand{\UtSrfrParentRefinementChangeExact}{+0.46}
\newcommand{\UtSrfrScratchRefinementChangeExact}{-0.35}
\newcommand{\UtSrfrScratchRangeExact}{0.92}

\renewcommand{\signalignerbf}{\textbf{SignAligner}\xspace}
\renewcommand{\scorebf}[1]{\textbf{#1}}
\renewcommand{\newpara}[1]{\vspace{1pt}\noindent\textbf{#1}}

\tikzset{every picture/.append style={scale=0.95,transform shape}}

\makeatletter
\def\bstctlcite#1{\@bsphack\@for\@citeb:=#1\do{%
  \edef\@citeb{\expandafter\@firstofone\@citeb}%
  \if@filesw\immediate\write\@auxout{\string\citation{\@citeb}}\fi}\@esphack}
\makeatother

\title{SignAligner: Dual-Granularity Span Generation for Sign Language Alignment}

\author{Jongbhin Woo$^{1}$\qquad Youngjoon Jang$^{2}$\qquad Joon Son Chung$^{1}$\\[3pt]
$^{1}$School of Electrical Engineering, KAIST, South Korea\\
$^{2}$Visual Geometry Group, University of Oxford, U.K.\\
{\tt\small jongbin.woo@kaist.ac.kr}}

\begin{document}
\bstctlcite{IEEEexample:BSTcontrol}
\twocolumn[{%
\renewcommand\twocolumn[1][]{#1}%
\vspace*{-0.3in}
\maketitle
\vspace{-1.2em}
\begin{center}
\captionsetup{type=figure,hypcap=false}
    \centering
    \begin{minipage}[c]{0.53\textwidth}
    \includegraphics[width=\linewidth]{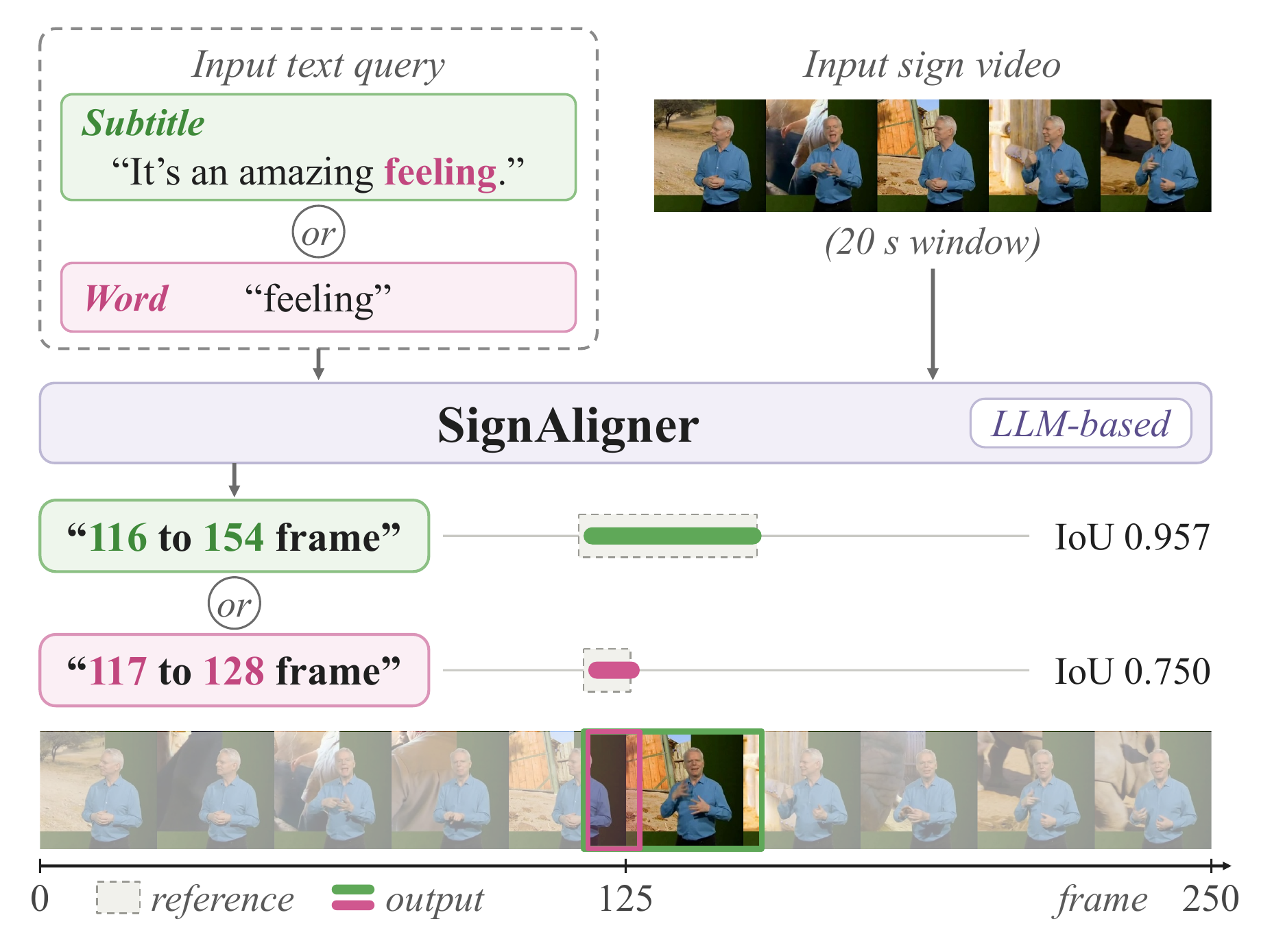}
    \end{minipage}\hfill
    \begin{minipage}[c]{0.44\textwidth}
    \centering
    \definecolor{tqSub}{HTML}{5FA858}\definecolor{tqSubText}{HTML}{3F8A3A}%
    \definecolor{tqSubLine}{HTML}{8CC084}%
    \definecolor{tqWord}{HTML}{D0578F}\definecolor{tqWordText}{HTML}{C2477F}%
    \definecolor{tqRefBar}{HTML}{D6D6D2}\definecolor{tqRefDark}{HTML}{5A5A5A}%
    \definecolor{tqInk}{HTML}{1F1F1F}\definecolor{tqNote}{HTML}{5A5A5A}%
    \definecolor{tqTick}{HTML}{444444}\definecolor{tqAxis}{HTML}{9A9A94}\definecolor{tqGrid}{HTML}{E9E9E5}%
    \begin{tikzpicture}[x=1pt,y=1pt,
      every node/.style={inner sep=0pt,font=\fontsize{8.3}{9.6}\selectfont,text=tqInk},
      note/.style={text=tqNote,font=\fontsize{8.3}{9.6}\selectfont\itshape},
      sub/.style={anchor=base west,font=\fontsize{8.3}{9.6}\selectfont},
      tick/.style={anchor=east,text=tqTick},
      cat/.style={anchor=base,text=tqNote},
      midlab/.style={anchor=base,text=tqSubText},
      ownlab/.style={anchor=base,text=#1,font=\fontsize{8.3}{9.6}\selectfont\bfseries},
      reflab/.style={anchor=base,text=tqNote},
      gain/.style={tqSubText,line width=0.65pt,-stealth},
      guide/.style={tqSubLine,line width=0.5pt,dash pattern=on 1.6pt off 1.2pt},
      tqax/.style={tqAxis,line width=0.5pt},
      tqgrid/.style={tqGrid,line width=0.45pt}]
    \useasboundingbox (0,0) rectangle (227,204.66);
    \node[anchor=base west,font=\fontsize{8.6}{10}\selectfont] at (0,192.8) {\textbf{\textit{\color{tqSubText}Subtitle}} alignment, F1@0.50};
    \node[sub] at (0,179.6) {Joint training (matched pair)};
    \node[sub] at (105,179.6) {Overlap-reward refinement};
    \node[note,anchor=base west] at (105,170.2) {other scoring setting};
    \node[cat] at (35,63.6) {Pair}; \node[cat] at (35,55.5) {SA-only};
    \node[midlab] at (84,63.6) {Pair}; \node[midlab] at (84,55.5) {joint};
    \node[cat] at (150,63.6) {Stage-1 model};
    \node[ownlab=tqSubText] at (204,63.6) {SignAligner};
    \foreach \ax in {14,119} {%
      \draw[tqax] (\ax,71.5) -- (\ax,68) (\ax,64.2) -- (\ax,61);
      \draw[tqTick,line width=0.6pt] (\ax-2.6,66.8) -- (\ax+2.6,69.4) (\ax-2.6,63.0) -- (\ax+2.6,65.6);}
    \begin{scope}[yshift=71.5pt]
    \begin{scope}[y=1.2pt,shift={(0,-25)}]
      \foreach \t in {50,75} {\draw[tqgrid] (14,\t) -- (96,\t) (119,\t) -- (218,\t);}
      \foreach \ax in {14,119} {%
        \draw[tqax] (\ax,100) -- (\ax,25);
        \foreach \t/\lab in {25/64,50/65,75/66,100/67} {\draw[tqax] (\ax,\t) -- (\ax-2,\t); \node[tick] at (\ax-4,\t) {\lab};}}
      \fill[tqRefBar] (25,25) rectangle (45,\JtOriginalSubtitleDotY);
      \node[reflab,yshift=3pt] at (35,\JtOriginalSubtitleDotY) {\JtOriginalSubtitleFOne};
      \fill[tqSubLine] (74,25) rectangle (94,\JtOriginalJointDotY);
      \node[midlab,yshift=3pt] at (84,\JtOriginalJointDotY) {\JtOriginalJointFOne};
      \draw[guide] (45,\JtOriginalSubtitleDotY) -- (64,\JtOriginalSubtitleDotY) (64,\JtOriginalJointDotY) -- (74,\JtOriginalJointDotY);
      \draw[gain] (64,\JtOriginalSubtitleDotY) -- node[anchor=east,xshift=-2.5pt,text=tqSubText,font=\fontsize{8.3}{9.6}\selectfont\bfseries] {\JtOriginalGain} (64,\JtOriginalJointDotY);
      \fill[tqRefBar] (140,25) rectangle (160,\CtParentDotY);
      \node[reflab,yshift=3pt] at (150,\CtParentDotY) {\CtParentFOneFifty};
      \node[note,anchor=base,yshift=24.2pt] at (150,\CtParentDotY) {longer};
      \node[note,anchor=base,yshift=14.6pt] at (150,\CtParentDotY) {joint run};
      \fill[tqSub] (194,25) rectangle (214,\RfRefinementMeanBarY);
      \node[ownlab=tqSubText,yshift=3pt] at (204,\RfRefinementMeanBarY) {\RfRefinementSeedMeanFOneZeroFifty};
      \draw[guide] (160,\CtParentDotY) -- (179,\CtParentDotY) (179,\RfRefinementMeanBarY) -- (194,\RfRefinementMeanBarY);
      \draw[gain] (179,\CtParentDotY) -- node[pos=0.66,anchor=east,xshift=-2.5pt,text=tqSubText,font=\fontsize{8.3}{9.6}\selectfont\bfseries] {\CtRefinementMinusParent} (179,\RfRefinementMeanBarY);
      \draw[tqax] (14,25) -- (96,25) (119,25) -- (218,25);
    \end{scope}
    \end{scope}
    \node[anchor=base west,font=\fontsize{8.6}{10}\selectfont] at (0,45) {\textbf{\textit{\color{tqWordText}Sign}} spotting, F1@0.50};
    \node[note,anchor=base west] at (104,45) {off-centre windows};
    \node[anchor=base east,text=tqNote] at (46,32.2) {Fixed centre};
    \node[ownlab=tqWordText,anchor=base east] at (46,21.7) {SignAligner};
    \begin{scope}[xshift=50pt]
    \begin{scope}[x=1.92pt,shift={(-25,0)}]
      \foreach \t in {62.5,100} {\draw[tqgrid] (\t,18.1) -- (\t,41);}
      \foreach \t/\lab in {25/0,62.5/20,100/40} {\draw[tqax] (\t,18.1) -- (\t,16.1); \node[anchor=base,text=tqTick] at (\t,8.7) {\lab};}
      \fill[tqRefBar] (25,31) rectangle ([shift={(2,0)}]\WlFixedCentreSquareY,39);
      \node[reflab,anchor=base west,xshift=3pt] at ([shift={(2,0)}]\WlFixedCentreSquareY,32.2) {\WlFixedCentreFOneHalf};
      \fill[tqWord] (25,20.5) rectangle (\WlNativeMeanBarY,28.5);
      \node[ownlab=tqWordText,anchor=base west,xshift=3pt] at (\WlNativeMeanBarY,21.7) {\WlNativeMeanFOneHalf};
      \draw[tqax] (25,18.1) -- (100,18.1) (25,18.1) -- (25,41);
    \end{scope}
    \end{scope}
    \end{tikzpicture}
    \end{minipage}
    \caption{\textbf{Prompt-conditioned span generation.}
    One generator answers subtitle and word queries with spans.
    Left: generated spans against references on one window. Right: joint training on a matched pair; refinement from the Stage-1 model under one fixed reconstruction setting; sign spotting against a control that always predicts the window centre.}
    \label{fig:teaser}

\end{center}
\vspace{0.6em}
}]

\begin{abstract}
Large sign-language corpora benefit from temporal labels at two scales: subtitle spans for subtitle alignment and word spans for sign spotting.
Sign spottings already locate words in time at corpus scale, but training subtitle alignment jointly on subtitle and word spans remains underexplored.
We present \signalignerbf, a prompt-conditioned visual-language span generator that maps a sign-language video and a subtitle-sentence or target-word query to a textual start/end frame interval.
It is first trained jointly on subtitle spans and word spans from sign spottings, then refined with overlap-based rewards while retaining auxiliary word examples.
On the subtitle test set of the British Broadcasting Corporation--Oxford British Sign Language (BOBSL) corpus, the refined model reaches \RfRefinementSeedMeanFOneZeroFifty{} F1@0.50 (F1 with a temporal intersection-over-union threshold of one half), the best among compared systems without external sign-video resources.
This overlap-reward refinement outperforms cross-entropy continuation from the same jointly trained model.
Joint training with word spans improves subtitle alignment in a pair of our generators matched in code, labels and maximum update budget. The advantage remains after both models of the pair are refined with overlap rewards.
For word queries, the refined model follows the queried word, not a fixed position in the video: when the word's sign lies away from the video centre, it is far more accurate than predictions that use position alone.
Together, these findings support one shared span generator refined with overlap rewards, aligning subtitles and producing word spans for human review.

\end{abstract}

\section{Introduction}

Large sign-language corpora need temporal annotation at two granularities. Subtitle alignment localises sentence-level subtitles in the signing stream \cite{bull2021aligning,jang2026deep,jang2025lostembeddings}, while sign spotting localises individual signs \cite{momeni2022automatic}. Both label types support corpus construction, retrieval, translation, and educational tools \cite{camgoz2018neural,camgoz2020sign,duarte2022signretrieval,cheng2023cico,tanzer2024youtubesl}. Subtitle timing pairs a subtitle with its video segment for translation \cite{camgoz2018neural,camgoz2020sign} and retrieval \cite{duarte2022signretrieval,cheng2023cico}. Word timing links a queried word to a local sign instance, which supports spotting-based annotation \cite{momeni2022automatic} and recognition from spoken-language supervision \cite{raude2024tale}. Both label types are costly to annotate manually.

Broadcast subtitles follow speech rather than signing, so aligning them by hand takes an expert about 10--15 hours per hour of video \cite{bull2021aligning}. Manually aligned subtitles are therefore scarce: they exist for only a small subset of episodes. Audio-aligned subtitles cover the larger corpus \cite{albanie2021bobsl}, and existing aligners retime them to the signing automatically \cite{bull2021aligning,jang2026deep}. Word-level temporal labels also exist at scale: dense spotting pipelines produce word-level sign spottings \cite{momeni2022automatic} that carry finer temporal information, though they are incomplete.

Current subtitle aligners learn sentence boundaries \cite{bull2021aligning,jang2026deep} or combine pretrained components \cite{jiang2025sea}. Spotting pipelines detect individual signs rather than align complete subtitle sentences \cite{momeni2020watch,varol2021read,momeni2022automatic}. Two problems remain. First, in framewise aligners, word timing stops supervising the predictor after pretraining. These aligners \cite{bull2021aligning,jang2026deep}, which label each frame as inside or outside the subtitle, use word labels mainly to pretrain their predictor and then train it on subtitles alone. Second, further supervised training on the scarce manual subtitle spans does not improve, on average, a model trained jointly on subtitle spans and sign spottings (Section~\ref{sec:refinement_results}). This holds under one fixed reconstruction setting, where reconstruction resolves per-subtitle spans into one episode timeline. Subtitle alignment is scored by span overlap against a threshold, which token-level cross-entropy does not encode.

For the first problem, we keep word timing in the training signal: we use the existing sign spottings to improve subtitle alignment by learning both tasks jointly. We introduce \signaligner, a prompt-conditioned span generator with shared parameters for subtitle and word queries. A subtitle sentence and a target word differ in query granularity, but both ask the model to identify a start/end frame interval. We therefore express subtitle alignment (\sa) and sign spotting (\ssp) as the same text-generation problem. Both query types share every trainable parameter and the answer format. The prompt identifies the granularity, and joint training (Stage~1) mixes subtitle and word examples. Sign spottings therefore directly train the predictor that answers subtitle queries. Subtitle spans cover whole sentences, whereas word spans built from sign spottings are short fixed-length windows anchored at each spotted sign. Joint training thus adds targets that subtitle spans alone do not supply. At inference, subtitle alignment uses subtitle queries alone, while word queries produce \emph{word spans}. Figure~\ref{fig:teaser} shows the shared generator answering a subtitle query and a word query on the same video. Each query is answered in a fixed window of the video, so a word span could come from the window position alone. We test whether these word spans follow the queried word instead.

For the second problem, overlap-reward refinement optimises a reward on the overlap of the generated interval with the reference, on the small manually aligned training subset of earlier alignment work \cite{bull2021aligning,jang2026deep}. For each query we sample candidate spans and score their overlap with the reference. A group relative policy optimisation (GRPO)-derived update \cite{shao2024deepseekmath} raises the probability of samples that score above their group mean, and a Kullback--Leibler term keeps the generator close to the jointly trained Stage-1 model. The Stage-1 model already answers both query types, so refinement adjusts the intervals of a working generator instead of relearning the task. Refinement keeps auxiliary word examples in its training stream (Stage~2).

\begin{itemize}
\item \textbf{Overlap-reward refinement improves subtitle alignment.} \signaligner reaches $\RfRefinementSeedMeanFOneZeroFifty$ F1@0.50 (Table~\ref{tab:sa-main}), the best among compared systems without external sign-video resources, and outperforms cross-entropy continuation from the same Stage-1 model (Section~\ref{sec:refinement_results}).
\item \textbf{Joint training with word spans improves our generator's subtitle alignment in a matched pair trained with the same code, labels and maximum update budget.} We introduce dual-granularity span training, which uses existing sign spottings as word-span supervision for subtitle alignment. One text-conditioned generator learns both tasks without task-specific heads. The gain is about one F1@0.50 point and remains after both models of the pair are refined with overlap rewards (Section~\ref{sec:joint_comparisons}).
\item \textbf{The refined generator localises query words beyond position shortcuts when the sign lies away from the window centre.} It scores far above position-only controls, while replacing the query word sharply lowers its score (Section~\ref{sec:word_level}).
\end{itemize}

\section{Related Work}

\newpara{Sign subtitle alignment.}
Aligning a subtitle means finding its boundaries in continuous signing.
Broadcast alignment must also absorb the offset between speech-timed subtitles and the signing. In our setting, an audio-aligned subtitle prior supplies the initial timing.
Other work takes temporal information from the signing itself. Mouthing cues enabled weak lexical supervision at scale in BSL-1K \cite{albanie2020bsl1k}. Automatic subtitle-unit segmentation \cite{bull2020automatic} and linguistically motivated sign-language segmentation \cite{moryossef2023linguistically} provide complementary approaches to identifying temporal boundaries.
On BOBSL, which introduced subtitle-alignment benchmarks \cite{albanie2021bobsl}, the Subtitle Aligner Transformer (SAT) corrects noisy timestamps \cite{bull2021aligning}. \satplus, an extension of the Subtitle Aligner Transformer, added BSL-grammar subtitle pre-processing, a selective alignment loss with negative video--subtitle pairs, and self-training \cite{jang2026deep}. SAT/\satplus pretrain a framewise aligner on sign spotting before subtitle training. SPOT-ALIGN studies sign retrieval through learned sign--text alignment \cite{duarte2022signretrieval}.
SEA \cite{jiang2025sea} combines pretrained sign segmentation, sign--text embeddings, and sequence-level dynamic programming for alignment across datasets. LiTFiE, concurrent work, generates start/end frame tokens with an added digit-wise L1 boundary loss, after pretraining on about \PubLitfiePretrainingHours{} hours from \bobsl and YouTube-SL-25 \cite{jang2025lostembeddings}. \signaligner trains one interval generator on word and subtitle spans, then applies overlap-reward refinement.

\newpara{Sign spotting and weakly supervised recognition.}
Mouthing cues, subtitles, and sign dictionaries \cite{momeni2020watch}, and attention-based localisation from weakly aligned subtitles \cite{varol2021read}, support automatic spotting pipelines that produce millions of sign spottings \cite{momeni2022automatic}. Continuous sign language recognition (CSLR) learns sign sequences without frame-level boundary labels, from sentence-level supervision \cite{cui2019iterative} or weak spoken-language supervision \cite{raude2024tale}. \signaligner reuses the mined word spans as a training signal for subtitle alignment: their temporal boundaries supervise the same predictor as the subtitle spans. Thus, existing dense annotation supports an additional task, despite the labels' incompleteness and mining bias.

\newpara{Multi-granularity sign-language learning.}
Recent sign language translation (SLT) methods combine pretrained models, adapters, and pseudo-labels \cite{chen2022simple,wong2024sign2gpt,gong2024signllm,jang2025context}. Unified SLT/\sa and sign-language understanding (SLU) pretraining share prompt-conditioned representations across tasks \cite{jang2025lostembeddings,li2025unisign}, and general video--text work models several granularities within one correspondence or grounding objective \cite{lin2024multigranularity,fang2024hlgt}. The multi-granularity video--text methods exploit hierarchy within the query or the video. \signaligner instead uses separately mined word spans as additional training targets for the predictor that answers subtitle queries.
LiTFiE couples subtitle alignment with translation, gaining $\PubLitfieMultitaskMinusSsaFOneFifty$ F1@0.50 over its model fine-tuned on alignment alone \cite{jang2025lostembeddings}. \signaligner couples subtitle alignment with sign spotting. For the \bobsl results, it uses only \bobsl sign video. The auxiliary targets are word spans, and both tasks predict the same temporal quantity at different query granularities.

\newpara{Generative temporal localisation.}
Discriminative temporal grounding covers cross-modal matching, boundary regression, span prediction, start--end maps, transformer detection, and the unified UniVTG framework \cite{gao2017tall,zhang2020span,zhang2020twodtan,lei2021qvhighlights,lin2023univtg,tang2022framewise,wang2022crossmodaldyn}. Conditional diffusion restores frame semantics \cite{liu2024conditional}. Other work uses cheaper video-level labels \cite{wang2022weakly} or single-point timestamps \cite{xu2023point,dong2026point}, while prompt tuning adapts pretrained models \cite{zeng2024mmpt}. Video LLMs such as VTimeLLM \cite{huang2024vtimellm} and TimeChat \cite{ren2024timechat} write timestamps as text, as \signaligner does. Others add dedicated time representations \cite{qian2024momentor,huang2024lita,wang2025groundedvideollm,guo2025trace}. Multimodal LLMs expand and clarify weakly supervised boundaries \cite{li2025etc}, enrich moment-retrieval queries \cite{liu2025contextmr}, or combine audio cues and shot-aware token compression \cite{yu2026smart}. Self-critical sequence training optimises a non-differentiable captioning metric by policy gradient, using the reward of the model's own greedy output as the baseline \cite{rennie2017selfcritical}. Overlap-reward refinement instead baselines each output against the mean reward of its sampled group. Time-R1 trains timestamp generation for temporal grounding in general video with GRPO, a timestamp-aware IoU reward and a reasoning-template reward \cite{wang2025timer1}. VideoChat-R1 fine-tunes a video multimodal LLM with rule-based spatio-temporal rewards on several tasks, including temporal grounding \cite{li2025videochatr1}. Neither Time-R1 nor VideoChat-R1 addresses sentence and word queries in one model, or the reconstruction of spans into subtitle timing. Reinforcement learning has also refined sign boundaries in continuous recognition \cite{wei2021semantic}. \signaligner generates spans as text and trains a shared predictor with sentence- and word-level span supervision.

\section{Method}

\subsection{Overview}

\begin{figure*}[t]
    \centering
    \includegraphics[width=0.86\linewidth]{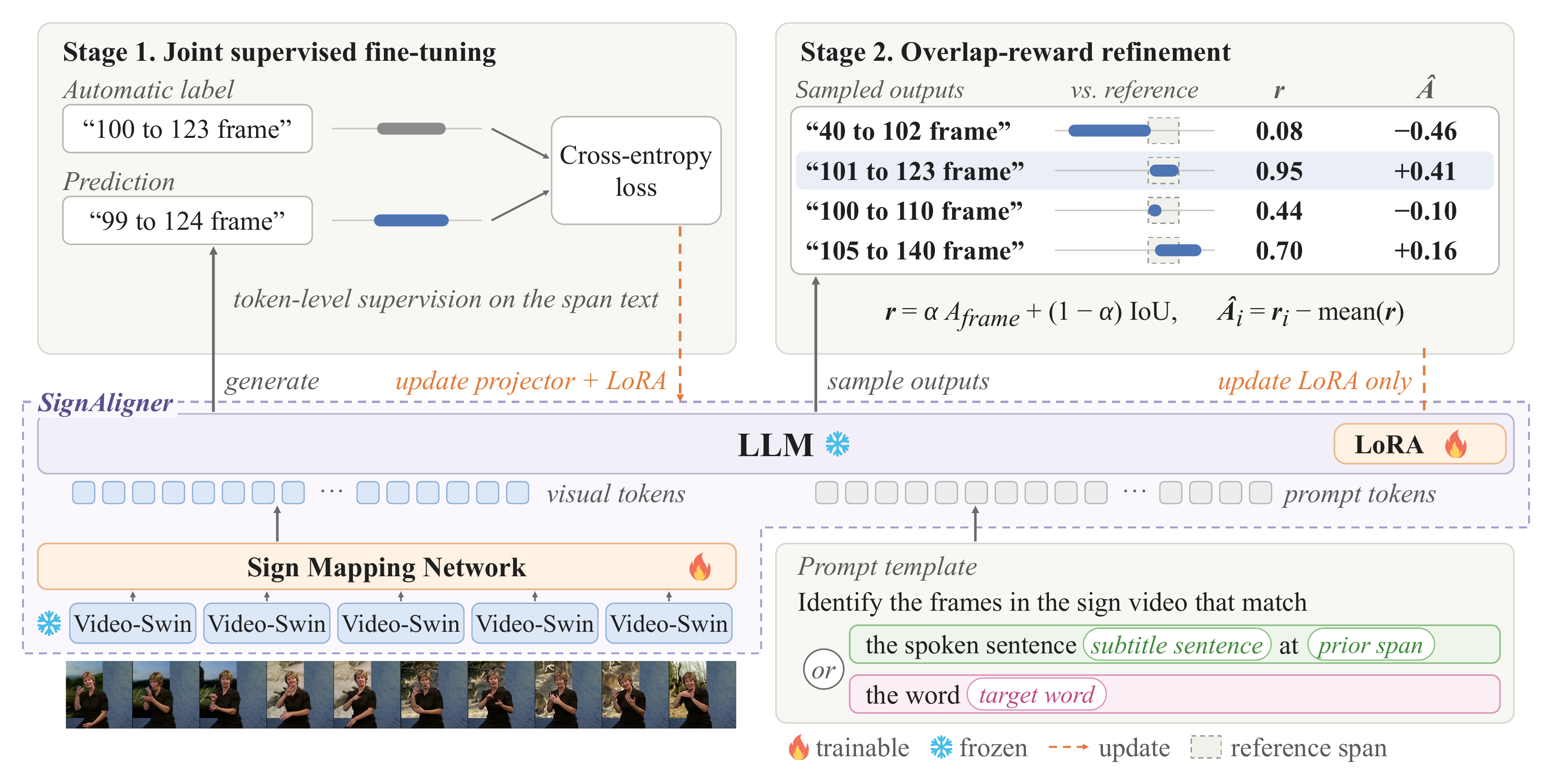}
    \caption{\textbf{Overview of \signaligner.}
    The prompt selects a subtitle or word query, and the decoder generates the span as text.
    Stage 1 trains on both span pools, and Stage 2 refines the model with overlap rewards.
    Drawn spans and rewards are illustrative, on the scale of Eq.~\eqref{eq:span_reward}.}
    \label{fig:overview}
\end{figure*}

\signaligner is one generator trained in two stages (Figure~\ref{fig:overview}). Stage~1 trains subtitle alignment and sign spotting jointly as text-conditioned span prediction. Stage~2 refines the resulting Stage-1 model with overlap rewards (Section~\ref{sec:method_stage2}). Both subtitle alignment and sign spotting ask for a start/end interval, so they share one output format: the span written as text, ``\{start\} to \{end\} frame''. For \sa, the subtitle query asks for the interval in which the sentence is signed. For \ssp, the target-word query asks for the corresponding sign instance. Both tasks share the visual encoder, projector, decoder, and inference span parser, but use task-specific prompts, supervision, and window selection. Parsed word spans go to sign-spotting evaluation. Subtitle spans undergo episode reconstruction (Section~\ref{sec:span_inference}).

Let $x$ be the features of a \JtWindowSeconds-second video window, $q$ a text query, and $s^*=(b^*,e^*)$ a reference start/end span in window-frame coordinates. The model decodes text $\hat{y}$, parsed into a span $\hat{s}=(\hat{b},\hat{e})$ in these coordinates and converted to the evaluation timeline:
\begin{equation}
    \hat{y}=\mathrm{decode}_\theta(x,q), \qquad \hat{s}=\mathrm{parse}(\hat{y}).
\end{equation}
The prompt, not a separate head, identifies the task. A subtitle query asks the model to ``Identify the frames in the sign video that match the spoken sentence `\{sentence\}' at `\{prior\}','' where the prior is the audio-aligned span; a word query asks for the frames that ``match the word `\{word\}'.'' Both answers use integer indices on the window's \JtFeatureGridFps\,fps feature grid (\JtWindowPositions{} positions). The inference span parser takes the first span written in this format; an output without one is unparsable.

\subsection{Architecture}

\newpara{Visual encoder.}
A Video-Swin encoder \cite{liu2022video}, kept frozen during \signaligner training, extracts dense features from overlapping windows. It is pretrained on \bobsl for isolated sign language recognition (ISLR) \cite{jang2025context}. These representations feed the shared generator.

\newpara{Projector and decoder.}
A lightweight MLP projector, the Sign Mapping Network in Figure~\ref{fig:overview}, maps these features into the embedding space of the decoder. The visual tokens enter the prompt through a dedicated sign-video slot of a frozen Llama 3.2 3B Instruct decoder \cite{meta2024llama32} adapted with LoRA \cite{hu2021lora}. In Stage 1, both subtitle and word examples update the shared projector and adapters. Stage 2 updates only the adapters.

\subsection{Two-Stage Training}

\newpara{Stage 1: dual-granularity supervised learning.}
Stage 1 trains the shared predictor with autoregressive cross-entropy over both pools:
\begin{equation}
    \mathcal{L}_{\mathrm{S1}}
    = \mathbb{E}_{\substack{(x,q,y)\sim\\ \mathcal{D}_{\mathrm{SA}}\cup\mathcal{D}_{\mathrm{SS}}}}
       \Big[-\textstyle\sum_t \log p_\theta(y_t \mid y_{<t}, x, q)\Big],
    \label{eq:stage1_loss}
\end{equation}
where $y$ is the target text of $s^*$. $\mathcal{D}_{\mathrm{SA}}$ comes from \satplus{} labels \cite{jang2026deep} and $\mathcal{D}_{\mathrm{SS}}$ from the automatic sign spottings of Momeni et al.~\cite{momeni2022automatic}, filtered by confidence (Section~\ref{sec:setup}).
Mini-batches are sampled uniformly at random from the union of both pools, with no task-specific loss multiplier or tuned ratio. Only positive word spans are used, so an unlabelled word is never treated as absent.

\newpara{Stage 2: overlap-reward refinement.}
\label{sec:method_stage2}
Stage 2 continues from the Stage-1 model on a mixed stream: the small manually aligned \sa training split plus confidence-filtered word spans from the manual training episodes as auxiliary word examples.
Stage 2 optimises the generated interval directly with an overlap-based reward rather than more supervised fine-tuning.
The update is a group-baselined policy gradient from the GRPO family \cite{shao2024deepseekmath,wang2025timer1}, rewarding overlap against manual subtitle references and automatic word references.
For a generated span $\hat{s}$ and reference span $s^*$, the reward combines a frame-aware term with temporal intersection-over-union:
\begin{equation}
    r_{\mathrm{span}} = \alpha\,A_{\mathrm{frame}}(\hat{s}, s^*) + (1-\alpha)\,\mathrm{IoU}(\hat{s}, s^*) .
    \label{eq:span_reward}
\end{equation}
The frame-aware term multiplies frame-mask accuracy over the window by reference-span recall,
\begin{equation}
    A_{\mathrm{frame}}(\hat{s}, s^*) = \frac{1}{W}\sum_{k=1}^{W}\mathbf{1}\big[m_k(\hat{s})=m_k(s^*)\big]\cdot\frac{|\hat{s}\cap s^*|}{|s^*|} ,
    \label{eq:frame_reward}
\end{equation}
where $W=\JtWindowPositions$ is the number of window positions, $m_k(s)$ indicates whether position $k$ lies in span $s$, and $|\cdot|$ counts positions.
Each prompt produces $G{=}\SuGenerationsPerPrompt$ sampled outputs $o_i$. Each output $o_i$ receives a reward $r_i$: a scaled $r_{\mathrm{span}}$, or a fixed reward if it is unparsable. The loss adds a KL term towards the frozen Stage-1 model:
\begin{equation}
    \mathcal{L}_{\mathrm{S2}} = -\frac{1}{N_{\mathcal{B}}}\sum_{i\in\mathcal{B}}\sum_{t=1}^{|o_i|}\hat{A}_i\,\ell_{i,t} + \beta\,\widehat{D}_{\mathrm{KL}} ,
    \label{eq:stage2_loss}
\end{equation}
where $\hat{A}_i=r_i-\frac{1}{G}\sum_j r_j$ is the advantage of $o_i$ among the outputs of its prompt and $\ell_{i,t}$ is the log-probability of its token $t$ at the sampling temperature. $|o_i|$ and $N_{\mathcal{B}}$ count the tokens before end-of-sequence in $o_i$ and in micro-batch $\mathcal{B}$, and $\widehat{D}_{\mathrm{KL}}$ is a per-token KL estimate on the sampled tokens. Advantages omit standard-deviation scaling, as in Dr.\ GRPO \cite{liu2025drgrpo}, and token-level normalisation follows DAPO \cite{yu2025dapo}.

\subsection{Span Inference and Subtitle Reconstruction}
\label{sec:span_inference}

\newpara{Word span inference.}
The model decodes each target word in a fixed \JtWindowSeconds-second window without its sentence in the prompt. For sign-spotting evaluation \cite{momeni2022automatic} on \senttest (defined in Section~\ref{sec:setup}), the window is centred on the manually aligned subtitle interval. For \signtest, it is placed at random so that it contains the annotated sign (Section~\ref{sec:word_level}). Parsed spans are converted to episode time and, lacking confidence scores, evaluated directly by overlap.

\newpara{Subtitle span inference.}
For \sa, each subtitle is decoded independently in a \JtWindowSeconds-second window centred on its audio-aligned prior (BOBSL's heuristically corrected audio timing plus a constant \JtAudioPriorOffset\,s offset). An unparsable output falls back to that prior. At inference, the manual reference never sets this window.

\newpara{Subtitle post-processing.}
Local predictions can overlap or violate episode order, whereas \sa requires non-overlapping, monotonic alignments, so we apply dynamic time warping (DTW) reconstruction following the BOBSL protocol \cite{bull2021aligning,jang2026deep}. Span generation emits no frame-level confidence map, so each predicted span $\hat{s}_j=(\hat{b}_j,\hat{e}_j)$ becomes a length-aware Gaussian curve with centre $\mu_j=(\hat{b}_j+\hat{e}_j)/2$ and width $\sigma_j=(\hat{e}_j-\hat{b}_j)/\sigma_{\mathrm{ratio}}$, where $\sigma_{\mathrm{ratio}}$ is a reconstruction setting. Its value at frame $f$, normalised by its peak, gives the cost
\begin{equation}
    C_{fj}=1-\frac{\mathcal{N}(f;\mu_j,\sigma_j)}{\max_{f'}\mathcal{N}(f';\mu_j,\sigma_j)},
\end{equation}
and DTW turns these costs into globally consistent intervals.

\section{Experiments}

\subsection{Experimental Setup}
\label{sec:setup}

\newpara{Datasets and splits.}
We train and test on \bobsl, a large corpus of sign-interpreted BBC broadcasts with aligned subtitles \cite{albanie2021bobsl}. Its subtitles follow the audio track, while signing is temporally offset from speech. The compared systems start from this audio prior, except that SEA starts from \satplus alignments. Table~\ref{tab:data} lists the data each stage trains on and the split each result uses.
\sentval and \senttest are the manually aligned sentence-level validation and public-test subsets of prior work \cite{albanie2021bobsl,bull2021aligning,jang2026deep}, and \signtest holds \bobsl's manually verified sign-level test annotations \cite{albanie2021bobsl,momeni2022automatic}. No automatic-label pool contains a \sentval, \senttest, or \signtest episode.
For transfer to another sign language, we use the WMT-SLT SRF corpus \cite{muller2022findings} of daily news interpreted live into Swiss German Sign Language (DSGS). We follow the episode split of SEA \cite{jiang2025sea}, whose episodes have subtitles both audio-aligned and manually aligned to the signing. Our SRF models train on the manually aligned subtitles of the training episodes and are tested on the test episodes, with video features from a different extractor than on \bobsl.

\newpara{Training data.}
Stage-1 word examples are the automatic, unvalidated sign spottings of Momeni et al.~\cite{momeni2022automatic} with confidence above \JtWordConfidence, which brings the word pool to the scale of the \satplus subtitle pool. Each spotting becomes a fixed-length span at its annotated time, with one length per cue type (mouthing, dictionary, or another cue of \cite{momeni2022automatic}).

\newpara{Implementation.}
Stage 1 trains the shared projector and LoRA adapters on the frozen decoder for \JtTrainingEpochs{} epochs on mixed subtitle and word examples. Stage 2 updates only LoRA for \RfMaximumRefinementOptimiserSteps{} steps with both prompt types and selects checkpoints on manual \sa validation. Inference uses deterministic decoding. Stage 1 trains \CtStageOneTrainableMillions M of the model's \CtTotalParameterBillions B parameters, and Stage 2 trains \CtLoraMillions M. Training the Stage-1 model took about \SuStageOneGPUHours{} GPU-hours, and each Stage-2 run about \RfApproximateGpuHoursPerRefinementSeed. On pre-extracted features and supplied priors, subtitle generation and DTW reconstruction run at a real-time factor of \UtCombinedRealTimeFactor, excluding model loading. The translator of Section~\ref{sec:slt_utility} is Llama 3.2 3B Instruct with a trainable MLP sign projector and LoRA, trained for five epochs. The decoder-free control of Section~\ref{sec:joint_comparisons} trains about \JtDecoderFreeTrainableMillions M parameters.

\begin{table}[t]
\centering
\caption{\textbf{Stage 1 trains on automatic labels; every test split is manual or verified.} Ep.: episodes. SA, SS: subtitles and word examples or queries (\bobsl counts after filtering and standard exclusions). Labels: \satplus subtitles~\cite{jang2026deep}, automatic spottings~\cite{momeni2022automatic} (auto.), manual alignment or verified signs. --: not counted.}
\label{tab:data}
\footnotesize
\setlength{\tabcolsep}{2.5pt}
\begin{tabular*}{\columnwidth}{@{\extracolsep{\fill}}lllrrr@{}}
\toprule
Split & Use & Labels & Ep. & SA & SS \\
\midrule
\multicolumn{6}{@{}l}{\textbf{\bobsl}} \\
Train (auto. labels) & Stage 1 & \satplus, auto. & \JtWordEpisodes & \JtSubtitleRows & \JtWordRows \\
Train (manual) & Stage 2 & Manual, auto. & \CtManualTrainingEpisodes & \CtManualSubtitleRows & \CtAuxiliaryWordRows \\
\sentval & Selection & Manual & \UtSentValidationEpisodes & \RfFilteredSentValSubtitleCount & -- \\
\senttest & SA, SS tests & Manual & \RfSentTestEpisodeCount & \RfFilteredSentTestSubtitleCount & -- \\
\signtest & SS test & Verified signs & \WlSignTestEpisodes & -- & \WlFilteredSignTestQueries \\
\midrule
\multicolumn{6}{@{}l}{\textbf{SRF (DSGS)}} \\
Test & Transfer & Manual & \UtSrfrTestEpisodes & \UtSrfrTestSubtitles & -- \\
\bottomrule
\end{tabular*}
\par\smallskip
\begin{minipage}{\columnwidth}
\footnotesize
\signtest queries cover \SuDistinctSignTestWords{} words; the released set has \SuReleasedVerifiedSigns{} signs from \SuReleasedVerifiedClasses{} classes.
\end{minipage}
\end{table}

\newpara{Evaluation.}
Subtitle alignment uses the unmodified \RfBOBSLFeatureEvaluatorFrameRateFps-fps BOBSL/SAT evaluator: per-subtitle spans are resolved into a non-overlapping episode timeline by DTW reconstruction and scored by F1 at an IoU threshold, with counts pooled over all subtitles of the split. F1@0.50 is primary. Frame accuracy and F1@0.10/0.25 are secondary. Reconstruction settings are selected on validation or reused from a related model. Independent word queries are scored on raw spans without DTW reconstruction, against same-word ground truth in the episode.
Paired \RfBootstrapPercentileIntervalLevel\% intervals come from \RfEpisodeClusterBootstrapResamples{} episode-cluster bootstrap replicates over the public-test episodes and measure variation across episodes for fixed outputs.

\begin{table}[t]
\centering
\caption{\textbf{\signaligner leads systems without external sign-video resources on \bobsl \senttest.} Bold marks the best score within each sign-video-resource block. $^{\dagger}$Mean of three runs.}
\label{tab:sa-main}
\footnotesize
\setlength{\tabcolsep}{2.5pt}
\begin{tabular}{@{}lrrrr@{}}
\toprule
\multirow{2}{*}{Method} & \multicolumn{3}{c}{F1@IoU} & \multirow{2}{*}{FA} \\
\cmidrule(lr){2-4}
 & 0.10 & 0.25 & 0.50 & \\
\midrule
\multicolumn{5}{@{}l}{\textbf{Audio priors}} \\
\addlinespace[1pt]
$S_{\mathrm{audio}}$ \cite{bull2021aligning,albanie2021bobsl} & \PubAudioFOneTen{} & \PubAudioFOneTwentyFive{} & \PubAudioFOneFifty{} & \PubAudioFrameAccuracy{} \\
$S_{\mathrm{audio}}^{+}$ \cite{bull2021aligning,albanie2021bobsl} & \PubCorrectedAudioFOneTen{} & \PubCorrectedAudioFOneTwentyFive{} & \PubCorrectedAudioFOneFifty{} & \PubCorrectedAudioFrameAccuracy{} \\
\cmidrule(lr){1-5}
\multicolumn{5}{@{}l}{\textbf{\bobsl-only sign-video training}} \\
\addlinespace[1pt]
SAT \cite{bull2021aligning} & \PubSatFOneTen{} & \PubSatFOneTwentyFive{} & \PubSatFOneFifty{} & \PubSatFrameAccuracy{} \\
\satplus \cite{jang2026deep} & \PubSatPlusFOneTen{} & \PubSatPlusFOneTwentyFive{} & \PubSatPlusFOneFifty{} & \PubSatPlusFrameAccuracy{} \\
\rowcolor{black!9}{\signaligner}$^{\dagger}$ & \scorebf{\RfRefinementFOneTenMean} & \scorebf{\RfRefinementFOneTwentyFiveMean} & \scorebf{\RfRefinementSeedMeanFOneZeroFifty} & \scorebf{\RfRefinementFrameAccuracyMean} \\
\cmidrule(lr){1-5}
\multicolumn{5}{@{}l}{\textbf{+ external sign-video resources}} \\
\addlinespace[1pt]
SEA \cite{jiang2025sea} + \satplus\ prior & -- & -- & \PubSeaBobslFOneFifty{} & -- \\
LiTFiE (SSA) \cite{jang2025lostembeddings} & \PubLitfieSsaFOneTen{} & \PubLitfieSsaFOneTwentyFive{} & \PubLitfieSsaFOneFifty{} & \PubLitfieSsaFrameAccuracy{} \\
LiTFiE (Multitask) \cite{jang2025lostembeddings} & \scorebf{\PubLitfieMultitaskFOneTen} & \scorebf{\PubLitfieMultitaskFOneTwentyFive} & \scorebf{\PubLitfieMultitaskFOneFifty} & \scorebf{\PubLitfieMultitaskFrameAccuracy} \\
\bottomrule
\end{tabular}
\par\smallskip
\begin{minipage}{\columnwidth}
\footnotesize
FA is frame accuracy, the percentage of correctly assigned frames. SSA is sign-to-subtitle alignment.
Evaluation uses \RfSentTestEpisodeCount{} episodes at \RfBOBSLFeatureEvaluatorFrameRateFps{} fps with method-specific reconstruction.
Baselines are published values \cite{jang2026deep,jiang2025sea}, including Table~6 of \cite{jang2025lostembeddings}. SEA reports F1@0.50 only.
\end{minipage}
\end{table}
\begin{figure}[t]
\centering
\includegraphics[width=\columnwidth]{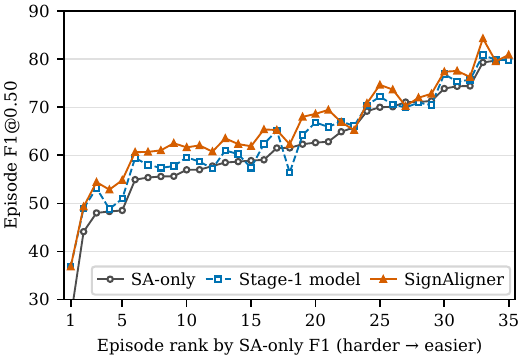}
\caption{\textbf{\signaligner exceeds \sa-only on most episodes.} Pooled F1@0.50: \sa-only \RfHistoricalSaOnlyTestFOneZeroFifty, Stage-1 model \RfHistoricalSatParentFOneAtOwnReconstruction, \signaligner (median run) \RfDiagnosticRefinedSeedFOneAtOneDecimal. \signaligner wins \RfFigureCaptionRefinementEpisodeWins{} episodes and loses three. \sa-only and the Stage-1 model use their own reconstruction settings.}
\label{fig:per-episode}
\end{figure}

\newpara{Baselines.}
For subtitle alignment, we compare against audio priors ($S_{\mathrm{audio}}$ and its heuristic correction $S_{\mathrm{audio}}^{+}$), SAT \cite{bull2021aligning}, and \satplus \cite{jang2026deep}, treating \satplus as the closest \bobsl-only baseline, compared by its published values. SAT and \satplus use fixed pretrained I3D sign-classification features with DTW reconstruction, and \satplus fine-tunes on manually aligned training subtitles, as our Stage~2 does. The SAT row is the value reported by \satplus.
Table~\ref{tab:sa-main} also includes SEA \cite{jiang2025sea} and LiTFiE \cite{jang2025lostembeddings}, grouped separately because they use external sign-video resources. SEA uses sign segmentation \cite{moryossef2023linguistically} pretrained on the Public DGS Corpus \cite{hanke2020extending} and SignCLIP \cite{jiang2024signclip} pretrained on SpreadtheSign \cite{spreadthesign}, then fine-tuned on \bobsl sign spottings. It tunes its alignment parameters on the training split and starts from \satplus subtitles. LiTFiE trains on \satplus pseudo-subtitles with pose and lip features.
For sign spotting, we list the dense-annotation pipelines of Momeni et al.~\cite{momeni2022automatic}, which mine \bobsl subtitle word candidates through mouthing, dictionary, attention, pseudo-label, exemplar, and novel-class cues.

\newpara{Our models.}
The Stage-1 model is our generator after joint Stage-1 training. \signaligner refines it in three runs. \signaligner rows marked $^{\dagger}$ report their mean, \RfRefinementSeedMeanFOneZeroFifty{} F1@0.50 in Table~\ref{tab:sa-main}. The \emph{median run}, the run with the median \senttest score (\RfCorrectedRefinementSeedFortyTwoPooledFOne, shown as \RfDiagnosticRefinedSeedFOneAtOneDecimal{} in Figure~\ref{fig:per-episode}), serves most output analyses. The \sa-only and \ssp-only models are trained in Stage 1 on subtitle or word queries alone. The \emph{matched pair} consists of two further Stage-1 models, trained separately with the same code, \satplus labels and maximum budget of \JtMaximumUpdates{} updates (Section~\ref{sec:joint_comparisons}). The pair joint model trains on subtitle and word queries, and the pair \sa-only model on subtitle queries alone. Both are shorter-trained than the Stage-1 model, so their scores are not directly comparable with \signaligner's.

\subsection{Subtitle Alignment}
\label{sec:subtitle_alignment}

\newpara{\signaligner leads systems without external sign-video resources.}
\signaligner's F1@0.50 exceeds the published \satplus result by $\CtRefinementMinusPublishedSatPlus$ and SEA by $\CtRefinementMinusPublishedSea$ \cite{jiang2025sea}. Against \satplus, the gains are in F1 at all three thresholds, while frame accuracy is nearly unchanged (\RfRefinementFrameAccuracyMean{} against \PubSatPlusFrameAccuracy).
LiTFiE \cite{jang2025lostembeddings} reaches a higher F1@0.50 after pretraining on about \PubLitfiePretrainingHours{} hours of \bobsl and YouTube-SL-25 video. Its model fine-tuned on alignment alone (SSA, sign-to-subtitle alignment) is close to \signaligner, and its multitask model also fine-tunes on translation (Table~\ref{tab:sa-main}).
Systems without external sign-video resources train on \bobsl sign video only, including pretraining of our Video-Swin encoder, while our decoder adds external language pretraining. The comparison with SAT and \satplus therefore also changes the visual features (Video-Swin rather than I3D). Sections~\ref{sec:refinement_results} and~\ref{sec:joint_comparisons} isolate refinement and joint training within our generator. Per episode, the median run exceeds the \sa-only model in all but three of the \RfSentTestEpisodeCount{} episodes (Figure~\ref{fig:per-episode}), so its gain over that model is broad.

\newpara{The lead over \satplus does not need reconstruction.}
Before DTW reconstruction, the median run's raw spans already score \RfDtwReconstructionRawSpanFOne{} F1@0.50, above \satplus with reconstruction (\PubSatPlusFOneFifty). None of these raw spans falls back to the prior. Reconstruction under the SAT/\satplus protocol raises the median run to \RfCorrectedRefinementSeedFortyTwoPooledFOne. Its median signed start and end offsets are $\RfMedianSignedStartOffsetSeconds$ and $\RfMedianSignedEndOffsetSeconds$\,s, a small early bias, and its end error grows for the longest subtitles.

\begin{figure}[t]
\centering
\includegraphics[width=\columnwidth]{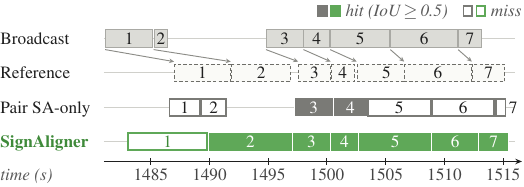}
\caption{\textbf{Broadcast subtitle times run ahead of the signing.} A selected excerpt of seven consecutive subtitles from one test episode. Arrows link each broadcast start to its reference start (dashed boxes). Bars: final spans of the pair \sa-only model (Section~\ref{sec:setup}) and \signaligner (median run); solid: hit.}
\label{fig:subtitle-cases}
\end{figure}

\newpara{Example spans show where \signaligner helps and where it fails.}
Figure~\ref{fig:subtitle-cases} follows seven consecutive subtitles of a selected test excerpt. Their broadcast times start before the signing, by several seconds for the first two. \signaligner aligns six of them, and the pair \sa-only model aligns two. Both models miss the first subtitle.

\begin{table}[t]
\centering
\caption{\textbf{Overlap-reward refinement improves the Stage-1 model; CE continuation does not on average.} All rows use one fixed reconstruction setting. $\Delta$ Stage-1 is the F1@0.50 change from the Stage-1 model. $^{\dagger}$Mean of three runs; other rows are single runs.}
\label{tab:continuation}
\footnotesize
\setlength{\tabcolsep}{4pt}
\renewcommand{\arraystretch}{1.12}
\settowidth{\dimen0}{F1@0.50}
\settowidth{\dimen2}{$\Delta$ Stage-1}\settowidth{\dimen4}{Prompts}
\edef\contlabelwidth{\the\dimexpr\columnwidth-\dimen0-\dimen2-\dimen4-6\tabcolsep-2pt\relax}
\begin{tabular}{@{}p{\contlabelwidth}rrr@{}}
\toprule
Recipe & Prompts & F1@0.50 & $\Delta$ Stage-1 \\
\midrule
Stage-1 model (no update) & \CtParentContinuationPrompts{} & \CtParentFOneFifty{} & -- \\
\cmidrule(lr){1-4}
\sa-only CE (one epoch)$^{\dagger}$ & \CtSaOnlyMaximumPrompts{} & $\CtSaOnlyCeMean$ & $\CtSaOnlyMinusParent$ \\
Mixed CE$^{\dagger}$ & \CtMixedMaximumPrompts{} & $\CtMixedCeMean$ & $\CtMixedCeMinusParent$ \\
Conservative CE: matched learning rate & \CtPilotOneMaximumPrompts{} & \CtPilotOneFOneFifty{} & $\CtPilotOneMinusParent$ \\
Conservative CE: matched schedule & \CtPilotTwoMaximumPrompts{} & \CtPilotTwoFOneFifty{} & $\CtPilotTwoMinusParent$ \\
\cmidrule(lr){1-4}
IoU-only overlap refinement & \CtIouMaximumPrompts{} & \CtIouOnlyFOneFifty{} & $\CtIouMinusParent$ \\
\rowcolor{black!9}\signaligner overlap refinement$^{\dagger}$ & \CtRefinementMaximumPrompts{} & \scorebf{\RfRefinementSeedMeanFOneZeroFifty} & $\CtRefinementMinusParent$ \\
\bottomrule
\end{tabular}
\par\smallskip
\begin{minipage}{\columnwidth}
\footnotesize
CE is cross-entropy. \sa-only CE trains on the manual subtitle examples; every other updated row adds the auxiliary word examples. Both conservative CE rows match refinement's learning rate; the matched-schedule row also matches its schedule, warmup and global batch. Prompts is the maximum prompt exposure, not the compute or the selected exposure.
\end{minipage}
\end{table}

\subsection{Overlap-Reward Refinement}
\label{sec:refinement_results}

The headline score might come from further training on the manual subtitles rather than from the overlap reward. We test this with cross-entropy continuation.

\newpara{Refinement improves alignment beyond CE continuation.}
Refinement and mixed cross-entropy (CE) continuation start from the same Stage-1 model and use the same manual subtitle and auxiliary word examples (Table~\ref{tab:continuation}). Under one fixed reconstruction setting, refinement's mean gain over mixed CE is \scorebf{\RfRefinementMinusMixedCEThreeSeedMeanFOneGainSigned} points. With reconstruction re-selected on validation, it stays $\CtReselectedRefinementMinusMixedCeFOneGain$ points ahead. One conservative CE variant matches refinement's learning rate. The other also matches its schedule, warmup and global batch (Table~\ref{tab:continuation}). Both stay behind refinement, as does one-epoch \sa-only CE.

\newpara{Refinement improves the Stage-1 model itself.}
Under the identical pipeline, refinement improves the Stage-1 model by $\CtRefinementMinusParent$ F1@0.50, while supervised recipes change it by $\CtPilotOneMinusParent$ to $\CtPilotTwoMinusParent$. Refinement's lead over CE therefore comes from refinement improving the Stage-1 model, not only from CE falling below it. IoU-only refinement, which keeps only the IoU term of Eq.~\eqref{eq:span_reward}, improves on it as well and stays above mixed CE. Its score is close to the median run's, so this comparison shows no advantage of the frame-aware term.

\subsection{Joint-Training Comparisons}
\label{sec:joint_comparisons}

Refinement starts from the Stage-1 model, which is trained jointly on subtitle and word spans. To test what word spans add, we compare the two models of the matched pair (Section~\ref{sec:setup}). We then test whether any advantage of joint training remains after refinement.

\begin{table}[t]
\centering
\caption{\textbf{Joint training scores higher for our generator and lower for the decoder-free control.} Scores are F1@0.50. Bold marks the best score in each comparison.}
\label{tab:task-ablation}
\label{tab:decoder-free}
\footnotesize
\setlength{\tabcolsep}{2pt}
\renewcommand{\arraystretch}{1.08}
\begin{tabular*}{\columnwidth}{@{\extracolsep{\fill}}lrr@{}}
\toprule
Generator / scoring & Pair \sa-only & Pair joint \\
\midrule
\multicolumn{3}{@{}l}{\textbf{\textit{Llama matched pair}}} \\
Llama, validation-selected checkpoints & \JtOriginalSubtitleFOne{} & \textbf{\JtOriginalJointFOne} \\
Same checkpoints, scored like the refined models & \JtMrBeforeSaFOneFifty{} & \textbf{\JtMrBeforeJointFOneFifty} \\
After overlap-reward refinement$^{\dagger}$ & \JtMrRefinedSaMeanFOneFifty{} & \textbf{\JtMrRefinedJointMeanFOneFifty} \\
\midrule
\multicolumn{3}{@{}l}{\textbf{\textit{Qwen pair}}} \\
Qwen3-4B, \JtQwenBudgetThousands k maximum updates & \JtQwenSubtitleFOne{} & \textbf{\JtQwenJointFOne} \\
\bottomrule
\end{tabular*}
\par\medskip
\setlength{\tabcolsep}{3pt}
\begin{tabular*}{\columnwidth}{@{\extracolsep{\fill}}lrr@{}}
\toprule
\multicolumn{3}{@{}l}{\textbf{\textit{Decoder-free control}}} \\
\midrule
Word fraction & F1@0.50 & $\Delta$ \\
\midrule
$\JtDecoderfreeSubtitleWordPercent\%$ (\sa-only) & \textbf{\JtDecoderfreeSubtitleFOne} & -- \\
$\JtDecoderfreeWordPercentSeven\%$ & \JtDecoderfreeSevenFOne{} & $\JtDecoderfreeSevenDelta$ \\
$\JtDecoderfreeWordPercentFourteen\%$ & \JtDecoderfreeFourteenFOne{} & $\JtDecoderfreeFourteenDelta$ \\
$\JtDecoderfreeWordPercentTwentySix\%$ & \JtDecoderfreeTwentySixFOne{} & $\JtDecoderfreeTwentySixDelta$ \\
$\JtDecoderfreeWordPercentFortyOne\%$ & \JtDecoderfreeFortyOneFOne{} & $\JtDecoderfreeFortyOneDelta$ \\
\bottomrule
\end{tabular*}
\par\smallskip
\begin{minipage}{\columnwidth}
\footnotesize
$^{\dagger}$Mean of three refinement runs. Both models are refined on subtitle and word examples.
The decoder-free control uses BERT text features and no LLM decoder, with the same visual features, episodes, ground truth, reconstruction and scorer.
Word fraction is the share of word examples in subtitle fine-tuning.
$\Delta$ is the change from \sa-only ($\JtDecoderfreeSubtitleWordPercent\%$ words).
\end{minipage}
\end{table}

\newpara{Joint training gains about one F1@0.50 point in the matched pair.}
With validation-selected checkpoints, the pair joint model scores higher than the pair \sa-only model, with a paired interval above zero (Table~\ref{tab:task-ablation}).
A Qwen3-4B pair also scores higher with joint training (Table~\ref{tab:task-ablation}), but its paired interval includes zero.

\newpara{Joint training's advantage persists through refinement.}
We refine both models of the matched pair with the same overlap-reward recipe, training data and evaluation. Under that evaluation, the pair joint model leads by $\JtMrDBefore$ F1@0.50 before refinement (second row of Table~\ref{tab:task-ablation}) and by $\JtMrMeanDAfter$ on average after it. It stays ahead in every paired refinement run. Joint training and overlap-reward refinement can therefore be used together in this matched pair.

\newpara{Joint training helps the generator, not the decoder-free control.}
In contrast to our generator, a framewise decoder-free control with BERT text features and no LLM decoder, on the same visual features, loses $\JtDecoderfreeSmallestLoss$--$\JtDecoderfreeLargestLoss$ F1@0.50 points with joint training at every tested word fraction (Table~\ref{tab:decoder-free}). This contrast motivates our LLM span generator.

\subsection{Sign Spotting}
\label{sec:word_level}

Joint training also produces word spans. Given a word as the query and a \JtWindowSeconds-s window, \signaligner generates the span in which that word is signed. \signtest asks whether this span matches a verified sign of the word (F1@0.50). \senttest asks whether the words of a subtitle are placed inside that subtitle, as scored by Momeni et al.~\cite{momeni2022automatic}.

\begin{figure}[t]
\centering
\includegraphics[width=\columnwidth]{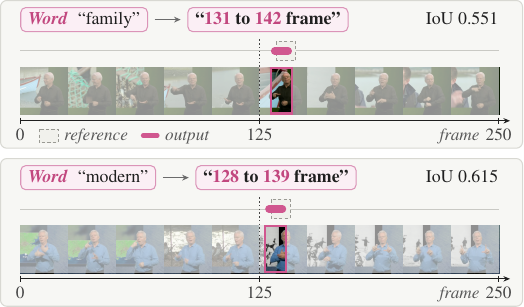}
\caption{\textbf{Off-centre word spans overlap the verified sign.} Two selected success cases from \signtest, each a word-only query in an off-centre window. The reference is a fixed \WlReferenceLengthSeconds-s span around the annotated time of the verified sign.}
\label{fig:word-cases}
\end{figure}
\begin{table}[t]
\centering
\caption{\textbf{Word spans score far above position-only controls off-centre.} Bold marks the best F1@0.50 within each \signtest block. $^{\dagger}$Mean of three runs.}
\label{tab:d1-main}
\label{tab:grounding-diagnostic}
\footnotesize
\setlength{\tabcolsep}{2pt}
\settowidth{\dimen0}{\scorebf{\WlNativeMeanFOneHalf}}
\settowidth{\dimen2}{F1@0.50}
\ifdim\dimen2>\dimen0 \dimen0=\dimen2\fi
\edef\wordlabelwidth{\the\dimexpr\columnwidth-\dimen0-2\tabcolsep\relax}
\begin{tabular}{@{}p{\wordlabelwidth}r@{}}
\toprule
\multicolumn{2}{@{}l}{\textbf{\signtest: \textit{off-centre localisation}}} \\
\midrule
System & F1@0.50 \\
\midrule
\rowcolor{black!9}{\signaligner}$^{\dagger}$ & \scorebf{\WlNativeMeanFOneHalf} \\
Fixed centre, \WlPositionControlLengthFrames{} frames & \WlFixedCentreFOneHalf{} \\
Uniform random, \WlPositionControlLengthFrames{} frames (mean) & \WlUniformRandomMeanFOneHalf{} \\
Whole window & \WlWholeWindowFOneHalf{} \\
\midrule
\multicolumn{2}{@{}l}{\textbf{Query word (median run)}} \\
Original word & \textbf{\WlNativeSeedFortyTwoFOneHalf} \\
Replaced word & \WlWrongQueryFOneHalf{} \\
\bottomrule
\end{tabular}
\par\medskip

\settowidth{\dimen0}{Recall}
\settowidth{\dimen4}{Coverage}
\edef\groundinglabelwidth{\the\dimexpr\columnwidth-\dimen0-\dimen4-4\tabcolsep\relax}
\begin{tabular}{@{}p{\groundinglabelwidth}rr@{}}
\toprule
\multicolumn{3}{@{}l}{\textbf{\senttest: \textit{subtitle-centred windows}}} \\
\midrule
System & Recall & Coverage \\
\midrule
\ssp-only & \WlSwsGroundingRecall{} & \WlSwsGroundingCoverage{} \\
Stage-1 model & \WlParentGroundingRecall{} & \WlParentGroundingCoverage{} \\
\rowcolor{black!9}\signaligner (median run) & \RfCorrectedSeedFortyTwoGroundingRecall{} & \RfCorrectedSeedFortyTwoGroundingCoverage{} \\
\midrule
\multicolumn{3}{@{}l}{\textit{Reference: published pipelines~\cite{momeni2022automatic}, their protocol}} \\
Base & \PubMomeniBaseSpotRecall{} & \PubMomeniBaseSpotCoverage{} \\
\quad + exemplars/novel classes & \PubMomeniExemplarsNovelSpotRecall{} & \PubMomeniExemplarsNovelSpotCoverage{} \\
\bottomrule
\end{tabular}
\par\smallskip
\begin{minipage}{\columnwidth}
\footnotesize
Position controls use neither text nor video.
\end{minipage}
\end{table}
\newpara{Word spans outperform position-only controls.}
On \signtest, each window is placed at random so that it contains the verified sign, usually off-centre, and the word alone is the query, without its sentence. \signaligner reaches a mean of $\WlNativeMeanFOneHalf$ F1@0.50, against at most $\WlPositionControlMaximumFOneHalf$ for controls that use only the window position (Table~\ref{tab:d1-main}). Replacing the query with a word with no verified sign in the window reduces the median run's score from \WlNativeSeedFortyTwoFOneHalf{} to \WlWrongQueryFOneHalf{}. In two off-centre windows, the output lands on the verified sign away from the window centre (Figure~\ref{fig:word-cases}). We place windows off-centre because centred windows favour position: in windows centred near each sign, a fixed centre prediction scores \WlCentredFixedCentreFOneHalf{} F1@0.50, above every model, including the median run (\SuHistoricalCentredSeedFOneFifty{}).

\newpara{On subtitle-centred windows, the \signaligner median run ranks above the \ssp-only and Stage-1 models.}
On \senttest, each subtitle word is decoded in a window centred on its subtitle and scored with the Recall and Coverage of Momeni et al.~\cite{momeni2022automatic}. Recall is the fraction of subtitle words whose predicted span lies at least half inside the subtitle, and Coverage is the fraction of subtitle frames covered by fixed-length spans at those words. Over \RfThresholdSweepEvaluatedSubtitles{} subtitles, the median run scores above the \ssp-only model and the Stage-1 model on both measures (Table~\ref{tab:grounding-diagnostic}). A fixed \WlPositionControlLengthFrames-frame span at the window centre again scores above our models (Recall \WlSentCentreControlRecall{}, Coverage \WlSentCentreControlCoverage{}). These rows therefore order our models but do not show localisation beyond window position. Table~\ref{tab:grounding-diagnostic} lists the published pipelines only for reference, under their own protocol.
\subsection{Downstream Subtitle-Crop Utility}
\label{sec:slt_utility}
Aligned subtitles define the video crops that translation models train on, so we test whether ours help. In an exploratory diagnostic, one translator recipe trains on crops from three sources, with the same training episodes and selection rule. Crops from the \signaligner median run give the highest CIDEr \cite{vedantam2015cider} and BLEURT-20 \cite{sellam2020bleurt}, slightly above \satplus crops (Table~\ref{tab:slt-utility}(a)). Both metrics use the manually aligned validation subtitles.

\begin{table}[t]
\centering
\caption{\textbf{Exploratory transfer and crop-utility checks.} (a) Translation trained with alternative subtitle crops. (b) Cross-lingual subtitle alignment. Bold marks the best score per metric in (a).}
\label{tab:srf}
\label{tab:slt-utility}
\footnotesize
\setlength{\tabcolsep}{3pt}
\textbf{(a) \bobsl subtitle-crop utility for translation training}\par\nobreak\smallskip
\begin{tabular*}{\columnwidth}{@{\extracolsep{\fill}}lrr@{}}
\toprule
Crop source & CIDEr & BLEURT ($\times\JtBleurtReportingScale$) \\
\midrule
SAT & \UtSatSelectedCropCider{} & \UtSatSelectedCropBleurt{} \\
\satplus & \UtSatplusSelectedCropCider{} & \UtSatplusSelectedCropBleurt{} \\
\signaligner & \scorebf{\UtSignalignerSelectedCropCider} & \scorebf{\UtSignalignerSelectedCropBleurt} \\
\bottomrule
\end{tabular*}
\par\smallskip
\begin{minipage}{\columnwidth}
\footnotesize
This is a \UtSentValidationEpisodes-episode validation diagnostic with one run per crop source. Both metrics use each run's CIDEr-maximising checkpoint.
\end{minipage}
\par\medskip
\textbf{(b) Cross-lingual subtitle alignment on SRF (DSGS)}\par\nobreak\smallskip
\begin{tabular*}{\columnwidth}{@{\extracolsep{\fill}}lrrrr@{}}
\toprule
\multirow{2}{*}{System} & \multicolumn{3}{c}{F1@IoU} & \multirow{2}{*}{FA} \\
\cmidrule(lr){2-4}
 & 0.10 & 0.25 & 0.50 & \\
\midrule
\multicolumn{5}{@{}l}{\textbf{Audio priors}} \\
\addlinespace[1pt]
$S_{\mathrm{audio}}$ & \UtSrfrAudioFOneTen{} & \UtSrfrAudioFOneTwentyFive{} & \UtSrfrAudioFOneFifty{} & \UtSrfrAudioFrameAccuracy{} \\
$S_{\mathrm{audio}}$ with fixed offsets & -- & -- & \PubSeaSrfOriginalPlusFOneFifty{} & -- \\
\cmidrule(lr){1-5}
\multicolumn{5}{@{}l}{\textbf{Published aligners}} \\
\addlinespace[1pt]
SAT, trained on SRF & -- & -- & \PubSeaSrfSatFOneFifty{} & -- \\
SEA & -- & -- & \PubSeaSrfFOneFifty{} & -- \\
\cmidrule(lr){1-5}
\multicolumn{5}{@{}l}{\textbf{Our generator, trained on SRF}} \\
\addlinespace[1pt]
From scratch$^{\dagger}$ & \UtSrfrScratchMeanFOneTen{} & \UtSrfrScratchMeanFOneTwentyFive{} & \UtSrfrScratchMeanFOneFifty{} & \UtSrfrScratchMeanFrameAccuracy{} \\
\quad + reward refinement on SRF & \UtSrfrScratchRefinedFOneTen{} & \UtSrfrScratchRefinedFOneTwentyFive{} & \UtSrfrScratchRefinedFOneFifty{} & \UtSrfrScratchRefinedFrameAccuracy{} \\
From the Stage-1 model & \UtSrfrParentFOneTen{} & \UtSrfrParentFOneTwentyFive{} & \UtSrfrParentFOneFifty{} & \UtSrfrParentFrameAccuracy{} \\
\quad + reward refinement on SRF & \UtSrfrParentRefinedFOneTen{} & \UtSrfrParentRefinedFOneTwentyFive{} & \UtSrfrParentRefinedFOneFifty{} & \UtSrfrParentRefinedFrameAccuracy{} \\
From \signaligner & \UtSrfrFinalFOneTen{} & \UtSrfrFinalFOneTwentyFive{} & \UtSrfrFinalFOneFifty{} & \UtSrfrFinalFrameAccuracy{} \\
From the pair \sa-only model & \UtSrfrMatchedSaFOneTen{} & \UtSrfrMatchedSaFOneTwentyFive{} & \UtSrfrMatchedSaFOneFifty{} & \UtSrfrMatchedSaFrameAccuracy{} \\
From the pair joint model & \UtSrfrMatchedJointFOneTen{} & \UtSrfrMatchedJointFOneTwentyFive{} & \UtSrfrMatchedJointFOneFifty{} & \UtSrfrMatchedJointFrameAccuracy{} \\
\bottomrule
\end{tabular*}
\par\smallskip
\begin{minipage}{\columnwidth}
\footnotesize
Evaluation uses the SEA split of \UtSrfrTestEpisodes{} episodes and \UtSrfrTestSubtitles{} subtitles. FA is frame accuracy. $^{\dagger}$Mean of three runs from scratch; every other row of ours is a single run. Reward refinement from scratch starts from the middle scratch run (\UtSrfrScratchMedianFOneFifty). The fixed-offset, SAT and SEA rows are published F1@0.50 values \cite{jiang2025sea}.
\end{minipage}
\end{table}

\subsection{Transfer to Another Sign Language}
\label{sec:srf_transfer}
Every result so far uses \bobsl, so we test whether \bobsl training transfers to another sign language. \bobsl initialisation improves F1@0.50 there. On SRF, an exploratory study trains our generator from scratch and from four \bobsl models: the Stage-1 model, \signaligner and the two models of the matched pair (Table~\ref{tab:srf}(b)). Each model transferred from \bobsl scores above every run from scratch at F1@0.50 (\UtSrfrParentFOneFifty--\UtSrfrMatchedJointFOneFifty{} against a from-scratch mean of \UtSrfrScratchMeanFOneFifty). Every transferred model also has higher frame accuracy than every run from scratch. At F1@0.25 and F1@0.10, every run from scratch scores higher than every transferred model.

Reward refinement on SRF changes F1@0.50 by $\UtSrfrParentRefinementChangeExact$ after transfer from the Stage-1 model and by $\UtSrfrScratchRefinementChangeExact$ from scratch, within the \UtSrfrScratchRangeExact-point spread of the scratch runs, so refinement is not shown to help on SRF.

Table~\ref{tab:srf}(b) also lists the audio priors and published aligners, whose visual components differ from ours \cite{jiang2025sea}. SRF has no word-level labels, so this study covers sentence-level alignment only.

\section{Limitations}
\label{sec:limitations}
First, English subtitle words serve as retrieval queries rather than BSL glosses, and word-level outputs are candidate annotations for human review, shaped by incomplete, mining-biased sign spottings.
Second, the reported metrics measure agreement with benchmark boundaries. Deaf-led or co-designed evaluation with Deaf or fluent BSL users is needed to assess linguistic adequacy before any deployed use \cite{desai2024systemic,bragg2021fate}.
Third, we study broadcast-interpreted BSL with audio-aligned subtitle priors. To our knowledge, \bobsl is the only public sign-language corpus that pairs subtitles automatically aligned to the signing with automatic word-level sign spottings at the scale of a thousand hours and millions of instances \cite{albanie2021bobsl,bull2021aligning,momeni2022automatic}. Corpora with manual sign-level time spans and sentence translations, such as the Public DGS Corpus and the BSL Corpus, are much smaller \cite{hanke2020extending,schembri2013building}. Conversational signing, settings without an audio-aligned prior and dual-granularity training in languages without comparable weak labels remain untested. We also have not tested reliance on \bobsl background or signer-identity cues, for example by masking non-signer regions.
Finally, \signaligner's refinement starts from one Stage-1 model. The joint-training comparison uses one training run per model, and the matched-pair refinement starts from those same models.

\section{Conclusion}

\signaligner reaches $\RfRefinementSeedMeanFOneZeroFifty$ F1@0.50, the best among compared systems without external sign-video resources. Its overlap-reward refinement exceeds cross-entropy continuation from the same Stage-1 model. By expressing subtitle alignment and sign spotting as the same text-conditioned span-prediction task, joint training brings existing sign spottings into the shared model. Joint training with word spans improves our generator's subtitle alignment, and the advantage persists through overlap-reward refinement in a matched pair. In off-centre sign-spotting windows, the refined generator localises query words far above position-only controls, and replacing the query word sharply lowers its score. In our generator on \bobsl, scarce manually aligned subtitles are better spent on an overlap reward than on further cross-entropy. Sign spottings that already exist at \bobsl scale also improve its subtitle alignment when both tasks are trained in one span generator. We hope these findings motivate alignment models that draw on the temporal labels a corpus already provides.

\section*{Acknowledgment}
We used Claude (Anthropic) to edit the manuscript text and to review code.
The authors reviewed and take full responsibility for all claims, results, and final manuscript text.

{\small
\bibliographystyle{IEEEtranN-refs}
\bibliography{references}
}

\end{document}